\documentclass[10pt,twocolumn,letterpaper]{article}

\usepackage{cvpr}
\usepackage{times}
\usepackage{epsfig}
\usepackage{graphicx}
\usepackage{amsmath}
\usepackage{amssymb}
\usepackage{algorithm2e}
\usepackage{color}
\usepackage{authblk}

\usepackage[pagebackref=true,breaklinks=true,letterpaper=true,colorlinks,bookmarks=false]{hyperref}

\cvprfinalcopy 

\def\cvprPaperID{2503} 

\ifcvprfinal\fi
\begin{document}

\title{Synthesizing 3D Shapes from Silhouette Image Collections using Multi-projection Generative Adversarial Networks}



\setlength{\affilsep}{0.1em}
\author[1,2]{\vspace{-0.33cm}Xiao Li}
\author[2]{Yue Dong}
\author[3]{Pieter Peers}
\author[2]{Xin Tong}
\affil[1]{University of Science and Technology of China}
\makeatletter
\renewcommand\AB@affilsepx{~~~~~~~ \protect\Affilfont}
\makeatother
\affil[2]{Microsoft Research Asia}
\affil[3]{College of William \& Mary\vspace{-0.3cm}}
\renewcommand\Authands{ and }


\maketitle

\definecolor{dblue}{rgb}{0.0,0.0,0.5}
\definecolor{dgreen}{rgb}{0.0,0.5,0.0}
\definecolor{dred}{rgb}{0.6,0.3,0.0}
\definecolor{dorange}{rgb}{0.6,0.25,0.0}
\definecolor{dyellow}{rgb}{0.5,0.5,0.0}
\newcommand{\NOTE}[1]{ \texttt{ \textbf{NOTE:} #1~ } }
\newcommand{\TODO}[1]{ \textcolor{blue}{\texttt{ \textbf{TODO:} #1~ }} }
\newcommand{\Yue}[1]{ \textcolor{dorange}{ (\emph{Yue}: #1)}  }
\newcommand{\Xin}[1]{ \textcolor{dgreen}{ (\emph{Xin}: #1)}  }
\newcommand{\PP}[1]{\textcolor{dblue}{ (\emph{PP}: #1)}}
\newcommand{\Xiao}[1]{\textcolor{dred}{ (#1)}}
\newcommand{\intd}{\,\text{d}}
\def\UNKNOWN{\textcolor{blue}{???}}
\newcommand{\ignore}[1]{}

\def\function#1{\mathbf{#1}}
\def\vector#1{\mathbf{#1}}

\def\p{\vector{x}}
\def\kd{k_d}
\def\ks{k_s}
\def\m{m}
\def\n{\vector{n}}
\def\w{\vector{\omega}}
\def\h{\vector{h}}
\def\fr{\function{f_r}}
\def\ior{\eta}
\def\NDF{\function{D}}
\def\G{\function{G}}
\def\F{\function{F}}

\def\L{\function{L}}
\def\X{\vector{X}}
\def\Z{\vector{Z}}
\def\V{\vector{V}}
\def\x{x}
\def\z{z}
\def\D{\function{D}}
\def\G{\function{G}}
\def\Y{\function{Y}}
\def\y{y}
\def\P{\function{P}}
\def\ZZ{\function{\Phi}}
\def\zz{\varphi}


\begin{abstract}

We present a new weakly supervised learning-based method for
generating novel category-specific 3D shapes from unoccluded image
collections.  Our method is weakly supervised and only requires
silhouette annotations from unoccluded, category-specific objects.
Our method does not require access to the object's 3D shape, multiple
observations per object from different views, intra-image
pixel-correspondences, or any view annotations.  Key to our method is
a novel multi-projection generative adversarial network (MP-GAN) that
trains a 3D shape generator to be consistent with multiple 2D
projections of the 3D shapes, and without direct access to these 3D
shapes. This is achieved through multiple discriminators that encode
the distribution of 2D projections of the 3D shapes seen from a
different views. Additionally, to determine the view information for
each silhouette image, we also train a view prediction network on
visualizations of 3D shapes synthesized by the generator. We
iteratively alternate between training the generator and training the
view prediction network.  We validate our multi-projection GAN on both
synthetic and real image datasets. Furthermore, we also show that
multi-projection GANs can aid in learning other high-dimensional
distributions from lower dimensional training datasets, such as
material-class specific spatially varying reflectance properties from
images.

\end{abstract}

\section{Introduction}
\label{sec:intro}

Learning to synthesize novel 3D shapes from a class of objects is a
challenging problem with many applications in computer vision, ranging
from single image 3D
reconstruction~\cite{choy20163d,fan2017point,sinha2017surfnet,Wu:2017:MarrNet,zhu2017rethinking},
to shape completion~\cite{Yang18}, to 3D shape
analysis~\cite{3dgan}. Typically, such methods are trained on
reference 3D shapes, or on images from different viewpoints with
labeled pixel-correspondences and/or with viewpoint
information. Creating such training sets for new object categories is
labor-intensive and cumbersome.  In this paper we propose a novel
weakly supervised method for learning to generate 3D shapes from
unoccluded silhouette image-collections without relying on reference
3D shapes, correspondences, or view annotations.

A key challenge in learning to synthesize 3D shapes from
image-collections gathered from uncontrolled sources is that we cannot
count on having multiple observations of the same object from
different viewpoints; we only have samples from the distributions of
images of the objects from different views. We overcome this practical
problem using a novel generative adversarial network (GAN)
architecture that learns the high dimensional distribution of 3D
shapes from multiple independently sampled low dimensional
distributions of silhouette images of the objects. The relation
between each low dimensional distribution of silhouette images and the
high dimensional 3D shapes is characterized by a
\emph{``projection''}, and the resulting \emph{multi-projection GAN}
architecture employs, for each projection, a different discriminator
that encodes the characteristics of the corresponding projection's
silhouette image distribution.  Intuitively, the proposed
multi-projection GAN learns the distribution of 3D shapes for which
the visualizations for each viewpoint matches the corresponding
(independent) distributions of the training silhouette images for that
same view.

A second key challenge is that the viewpoint information is often not
available for the training images that are gathered ``in the wild''.
To address this challenge, we utilize a view prediction network for
inferring the view information for each of the training images.
Ideally, the view prediction network should be trained by the types of
3D shapes synthesized by our generator. However, this creates a
circular dependency, because the 3D shape generator needs the view
prediction network to train from silhouette images.  We resolve this
dilemma by jointly training both networks in an iterative alternating
fashion, resulting in a 3D shape learning pipeline that is robust to
non-uniform viewpoint distributions over the training data (i.e., we
make no assumptions on the viewpoint distribution).

Our method facilitates training of a 3D shape generator from weakly
supervised real-world image collections with silhouette annotations,
thereby greatly reducing the cost for learning a 3D shape generator
for new object categories.  We demonstrate the strengths of our
multi-projection GAN with a jointly trained view prediction network on
both synthetic datasets and real-world silhouette image-collections.
Furthermore, we demonstrate that multi-projection GANs can be
generalized to aid in learning other types of high-dimensional
(non-image) distributions from low-dimensional image observations such
as learning the distribution of spatially-varying material reflectance
parameters.

\section{Related Work}
\label{sec:related}

In the past few years, various deep learning based methods have been
proposed for 3D shape generation. These methods can be roughly
categorized in three classes. A first class of methods relies on a
large set of reference 3D shapes for training (e.g., for generating 3D
voxel shapes with a VAE-GAN~\cite{3dgan}, for shape
completion~\cite{Yang18}, or for 3D reconstruction from a 2D
image~\cite{choy20163d,fan2017point,girdhar2016learning,jiang2018gal,sinha2017surfnet,Wu:2017:MarrNet,zhu2017rethinking}). Of
particular note is the method of Zhu~\etal~\cite{zhu2017rethinking}
that jointly reconstructs a single 3D voxel shape and camera pose.
A second class of methods foregoes the need for reference 3D shapes,
and uses annotated images with correspondences~\cite{cmrKanazawa18} or
exploit consistency over different views of
objects~\cite{mvcTulsiani18,Tulsiani:2017:MSS,Yan:2016:PTN}. Of
particular note is the method of Tulsiani~\etal~\cite{mvcTulsiani18}
who predict both shape and viewpoint.
Closest related to the proposed method is a third class of methods
that learns the 3D shape distribution from unannotated images.
Gadelha~\etal\cite{Gadelha2017} train a voxel GAN using a single
discriminator with 2D silhouette images from pre-defined discrete
 view distributions. Henderson~\etal~\cite{henderson18bmvc} utilize a
variational auto-encoder (VAE) for generating 3D meshes from
unannotated images with uniform distributed viewpoints.  In contrast
to Gadelha~\etal, we use multiple discriminators for each of the
different views, resulting in a better generation quality. Unlike
Henderson~\etal, we build on GANs as opposed the VAEs. More
importantly, we do not require a uniform viewpoint distribution over
the training dataset, which is difficult to enforce in internet-mined
datasets.  As we will show in~\autoref{sec:results}, for similar shape
representations, the proposed method produces higher quality results
compared to both prior methods.

Recently, Bora~\etal\cite{bora2018ambientgan} generalized the method
of Gadelha~\etal, and showed that a generative model (for images) can
be trained from different types of lossy projections. However,
Bora~\etal use a single discriminator, resulting in a suboptimal
synthesis for projections with different distributions. To handle
multiple projection distributions, our method relies on multiple
discriminators, i.e., one for each projection.  We are not the first
to consider combining multiple discriminators to train
GANs~\cite{Durugkar:2016:GMA,Neyshabur:2017:SGT,kanazawa2018end}.
However, all these prior methods require access to the high
dimensional training data, and focus on improving either the
efficiency or the stability of the training process.  In contrast, we
do not have access to the high-dimensional distributions (i.e., 3D
shapes), and train it directly from the lower-dimensional projections
(i.e., images).

\section{Shape Distributions from Silhouette Images}
\label{sec:shape}

We aim to learn a 3D shape distribution $\X$, in the form of a
generator, for a class of objects from a collection of uncorresponded
silhouette images that follows a distribution $\Y$. Our solution
(\autoref{fig:overview}) consists of two key components: a novel
multi-projection GAN (\autoref{sec:shape_gan_overview}) that learns
the 3D distribution $\X$ using multiple discriminators that ensure
that the corresponding 2D ``projections'' from the 3D distribution
follow the distribution of the training silhouette images, and a view
prediction network (\autoref{sec:shape_view_estimation}) for
estimating the viewpoints of the input 2D silhouette training
images. The training of both networks depends on the availability of
the other. We will therefore first introduce multi-projection GANs
assuming that estimates of the viewpoints are available. Next, we
introduce our view prediction network and the joint training strategy
for training both networks iteratively.

\subsection{Multi-Projection GAN}
\label{sec:shape_gan_overview}

\paragraph{GAN overview}
A Generative Adversarial Network
(GAN)~\cite{Arjovsky:2017:WGAN,Goodfellow2014,Radford:2015:URL}
consists of a generator network $\G$ and a discriminator network
$\D$. The generator $\G$ takes as input a vector of latent variables
$\z$ sampled from uniformly distributed noise $\Z$ and generates
samples from a learned distribution $\X$. The discriminator $\D$
judges whether a sample belongs to the distribution $\X$ or
not. Training both networks is performed in competition, until the
discriminator cannot distinguish the generated samples from the data
distribution. The loss for the discriminator $\D$ is defined as:
\begin{equation}
  \L_\D(\X, \Z) = \sum_{\x \sim \X} \log(\D(\x)) + \sum_{\z \sim \Z} \log(1 - \D(\G(\z))),
\end{equation}
and the loss for the generator is defined as:
\begin{equation}
  \L_\G(\Z) = \sum_{\z \sim \Z} \log(\D(\G(\z))).
\end{equation}
In the case of 3D shape generation, we represent each object in $\X$
by a binary 3D voxel grid.

\paragraph{Projection}
In order to train a 3D shape GAN, a large collection of exemplars from
the 3D shape distribution $\X$ are needed. However, such collections
only exist for a few object classes. In our case, we only have
silhouette images of the class of objects that follow a distribution
$\Y$. The dimensionality of $\Y$ (2D images) is lower than the
dimensionality of $\X$ (3D voxel grid). We assume that the relation
between the low dimensional distribution $\Y$ and the high dimensional
distribution $\X$ can be modeled by a differentiable (potentially
non-linear) \emph{projection}:
\begin{equation}
  \Y = \P(\X, \ZZ),
\end{equation}
where $\ZZ$ are the latent parameters of the projection (e.g., the
intrinsic and extrinsic camera parameters), which model any external
factors unrelated to the target distribution $\X$.

For learning 3D shape distributions, the projection $\P$ generates a
silhouette image from a 3D voxel shape. Practically, we follow the
method of Tulsiani~\etal.~\cite{Tulsiani:2017:MSS} exactly: Given a 3D
voxel shape and viewpoint, we first compute a ray intersection
probability for each voxel using ray-casting. Next, the silhouette is
computed as the expected value of the intersection probability along
the $z$ axis.

\begin{figure}[t]
	\begin{center}
		\includegraphics[width=0.9\linewidth]{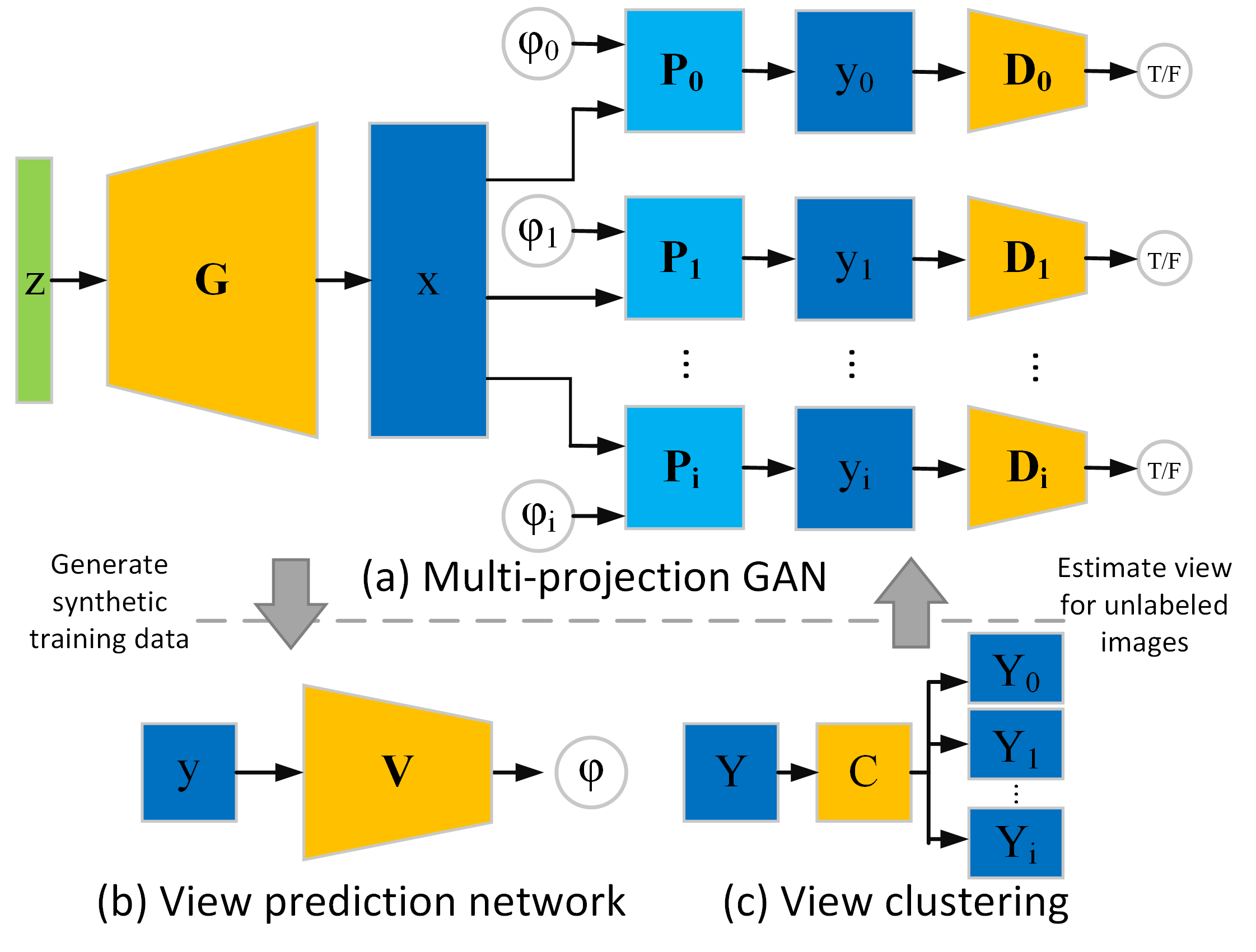}
		\caption{ \textbf{(a)} A multi-projection GAN consists
                  of a generator $\G$ that takes as input a vector of
                  latent variables $\z$ sampled from a uniform
                  distribution $\Z$, and generates samples from the
                  target distribution $\X$.  The generator $\G$ is
                  trained in competition with multiple discriminators
                  $\D_i$ that assess if the projection $\P_i$ of a
                  sample belongs to the projected target distribution
                  $\Y_i$. The projection $\P_i$ relates a high
                  dimensional sample from $\X$ (e.g., voxel shape) to
                  a lower-dimensional sample $\Y_i$ (e.g., silhouette
                  image), and which can feature its own latent
                  parameters ($\zz_i$) to model parameters independent
                  of $\X$ (e.g., viewpoint variations).  \textbf{(b)}
                  The view prediction network $\V$ estimates the
                  viewpoints $\varphi$ from a silhouette image $\Y$,
                  that are subsequently clustered and merged,
                  \textbf{(c)} and assigned to the training set of the
                  corresponding discriminator.  We iteratively
                  alternate between training both networks.}
		\label{fig:overview}
	\end{center}
\vspace{-0.3cm}
\end{figure}

\paragraph{Multi-projection GAN}
Generally, unless the high dimensional distribution $\X$ occupies a
low dimensional embedding, a single low dimensional projection will
incur a loss of information, and thus cannot unambiguously determine
the target distribution $\X$.  For example, without additional priors,
we cannot reconstruct the distribution of 3D shapes from silhouettes
from a single view.  Even when considering multiple views (modeled
through the latent parameters of the projection $\ZZ$), a single
discriminator is unlikely to be able to model the joint distribution
over all views with sufficient accuracy.  We therefore consider
multiple projections $\P_i$ with corresponding projected sample
distributions $\Y_i$ and associated discriminators $\D_i$.  We define
the combined loss function for each discriminator as:
\begin{align}
  \L_{\D_i}(\Y_i, \Z) &= \sum_{\y \sim \Y_i} \log(\D_i(\y)) + \\
  					  &\sum_{\z \sim \Z, \zz \sim \ZZ_i} \log(1 - \D_i(\P_i(\G(\z), \zz))).
\label{eq:d}
\end{align}
Similarly, we define the loss for the generator as:
\begin{equation}
  \L_\G(\Z) = \sum_{i} \sum_{\z \sim \Z, \zz \sim \ZZ_i} \log(\D_i(\P_i(\G(\z), \zz))),
\label{eq:g}
\end{equation}
where we randomly draw the latent variables $\z$ from $\Z$ , and the
latent projection parameters $\zz$ from a distribution $\ZZ_i$. In the
case of 3D shape generation from silhouette images, the distribution
$\ZZ_i$ encompasses the set of viewpoints that project to similar
silhouettes (i.e., to account for errors in the view
calibration). \autoref{fig:overview} (a) overviews our 3D shape
generation multi-projection GAN, named \emph{MP-GAN}.
%

\paragraph{Independence of $\Y_i$}
An advantage of our multi-projection approach is that the loss
function for the i-th discriminator (\autoref{eq:d}) only depends on
samples drawn from $\Y_i$.  Hence, each discriminator can be trained
from independently draw samples from each $\Y_i$. Consequently, the
samples for the different discriminators do not need to correspond to
the same objects or latent projection parameters.  The choice of which
samples to draw for the training for each discriminator does not
affect the loss function of the generator either. From a practical
perspective, this allows us to train a 3D shape generator with image
collections of different objects from different viewpoints, without
requiring any explicit correspondences.

\paragraph{Distinct Discriminator Requirement}
We also observe that a good choice of projections for training
multi-projection GANs are those whose projected data distributions are
significantly different.  Indeed, if two projections $\P_a(\X, \ZZ_a)$
and $\P_b(\X, \ZZ_b)$ are similar to each other, then the
corresponding discriminators will learn the same distribution. Hence,
according to~\autoref{eq:d}, the two discriminators will similarly
affect the training of the generator. Therefore, the two similar
projections should act as a single projection:
\begin{equation}
\P(\X, \ZZ) = \P_a(\X, \ZZ_a) \cup \P_b(\X, \ZZ_b), \quad \ZZ = \ZZ_a \cup \ZZ_b.
\label{eq:merge_d}
\end{equation}
Concretely, for 3D shape modeling, a large difference in viewpoint can
yield a significantly different silhouette image distribution. Hence,
different discriminators are needed for large viewpoint differences.
In contrast, small viewpoint changes produce very similar silhouette
distributions, and thus we can combine discriminators of similar views
together, and model the viewpoint variations by the latent projection
parameters $\ZZ$.  As a result, a multi-projection GAN for 3D shapes
does not require perfect viewpoint estimates; we will exploit this
property when training the generator and viewpoint predictor
(\autoref{sec:shape_view_estimation}).

\subsection{View Prediction and Clustering}
\label{sec:shape_view_estimation}
The multiple discriminators $\D_i$ and projections $\P_i$ in
\emph{MP-GAN} require knowledge about which silhouette images belong
to which distribution $\Y_i$ as well as the corresponding viewpoints
modeled by the latent projection parameters $\zz \sim \ZZ_i$.  Prior
work in deep viewpoint estimation relied on labeled training data or
synthetically rendered images from known 3D shape
collections~\cite{Massa:2016:CMC,Su:2015:RCV,WU:2016:SII}, or on
multi-view correspondences~\cite{mvcTulsiani18}. None of these methods
are directly applicable to our input training dataset.

To estimate the viewpoints, we utilize a view prediction network
trained on a large number of reference silhouette images (with
viewpoint) obtained by projecting 3D shapes \emph{synthesized} by the
generator $\G$.  For robustness, we discretize the space of possible
viewpoints into $16$ predefined \emph{``view-bins''} and treat the
view prediction as a classification problem that outputs a vector of
view probabilities.

\paragraph{Viewpoint Clustering}
Ideally, each view-bin corresponds to a projection and an associated
discriminator (\autoref{eq:d}). The silhouette images assigned to the
view-bin serve as a training set for the respective
discriminator. However, depending on the viewpoint distribution over
the training data, not all of the view-bins will contain a sufficient
number of training images to train a discriminator. Therefore, instead
of using all view-bins, we exploit the observation expressed
in~\autoref{eq:merge_d}, and merge nearby views to a
pre-determined fixed number of view clusters.

Practically, we perform a K-means clustering on the predicted
discretized view probability vectors of the training silhouette
images. We use the view probability vectors instead of the estimated
viewpoints to better handle ambiguous cases (e.g., front and back view
produce identical silhouettes). The result will be a set of viewpoint
distributions (as cluster centers) and a unique cluster \emph{id} for
each training image. We directly use the estimated viewpoints of the
silhouettes images assigned to the i$^{th}$ cluster as the latent
projection viewpoint distribution $\ZZ_i$ to ensure that the
(synthesized) projected distribution follows the intrinsic
distribution of the training silhouette data.

\begin{figure}[t]
	\begin{center}
		\includegraphics[width=\linewidth]{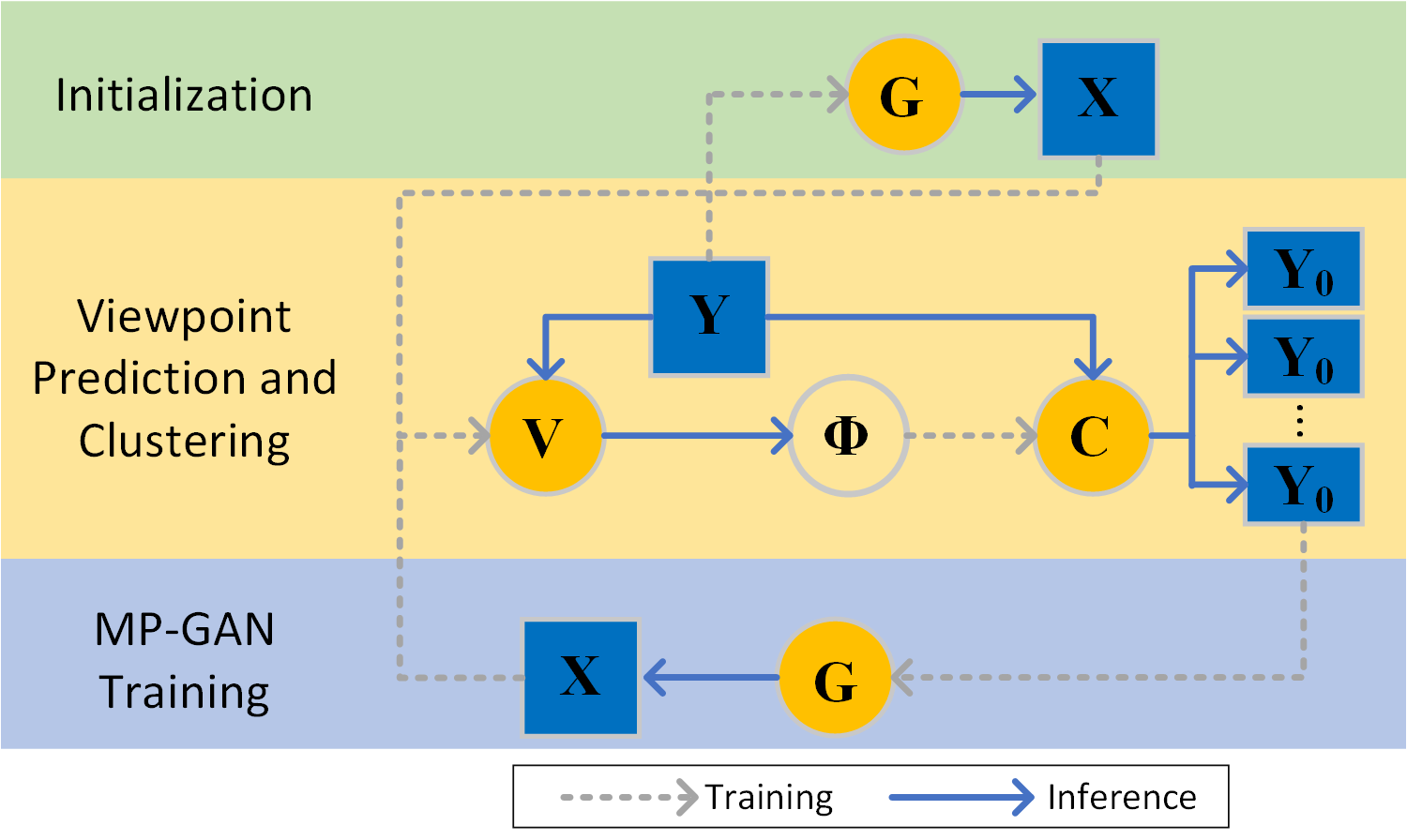}
		\caption{Overview of \emph{VP-MP-GAN}
                  training. Starting from a single discriminator GAN,
                  we iteratively alternate between training a
                  viewpoint classifier (based on training images
                  produced by the generator $\G$), and training
                  \emph{MP-GAN} (based on predicted viewpoints for the
                  silhouette images using the viewpoint classifier
                  $\V$).}
		\label{fig:auto_md_gan}
	\end{center}
\vspace{-0.3cm}
\end{figure}

\paragraph{Joint Training}
Both the multi-projection GAN and the viewpoint prediction classifier
require the other for training (i.e., multi-projection GAN requires
viewpoint estimates, and the view prediction classifier requires
GAN-generated 3D shapes for training). We resolve this conundrum by
jointly training both in an iterative fashion where we alternate
between training one network while keeping the other fixed.  We
bootstrap this iterative joint approach by training an initial
generator using a single-discriminator 3D shape GAN assuming uniformly
distributed viewpoints for training the discriminator. We will refer
to the multi-projection GAN trained with joint viewpoint predictions
as \emph{VP-MP-GAN}. \autoref{fig:auto_md_gan} summarizes our full
pipeline.

\section{Implementation and Training}
\label{sec:shape_implementation}

\paragraph{Network Structure}
We follow the 3D voxel generator network structure of
Wu~\etal~\cite{3dgan} which takes a input vector sampled from $N(0,1)$
and outputs a $64 \times 64 \times 64$ voxel grid via a sequence of 3D
convolution and upsample layers.  We employ batch normalization after
each convolution and upsample layer and use the ReLU activation
function. The discriminators take a $64 \times 64$ binary silhouette
image as input, and output a single scalar probability value. Each
discriminator contains $4$ convolutional blocks with a stride of $2$,
followed by a single fully-connected layer. We share the first
convolutional layers among the different discriminators, and use
spectrum normalization~\cite{miyato2018spectral} for each layer with
the LeakyReLU activation function.  The view prediction network shares
the same convolutional structure as the discriminators, but with using
batch normalization instead of spectrum normalization, and outputting
the probability for the $16$ discretized view-bins.  We refer to the
supplemental material for the detailed network structures.

\paragraph{Training Data}
We demonstrate our method on four different datasets: a synthetic
dataset for validation and ablation study, and three different
real-world datasets.  The synthetic dataset consists of $40,000$
rendered silhouette images for $6,\!000$ 3D shapes of chairs from the
ShapeNet dataset~\cite{Chang:2015:SNA}. All silhouettes are rendered
from viewpoints with no elevation angle and uniformly distributed
azimuth angles in $[0, 2\pi]$.  The three real-world datasets are:
\mbox{$\sim\!\!22,\!000$} chair images mined from internet image
repositories and \emph{Pix3D}~\cite{pix3d}, a dataset of
\mbox{$\sim\!\!36,\!000$} car images mined from internet image
repositories plus the Stanford car
dataset~\cite{KrauseStarkDengFei-Fei_3DRR2013}, and a dataset of
\mbox{$\sim\!\!12,\!000$} bird images
(\emph{CUB-Birds-200-2011})~\cite{WahCUB_200_2011}. For \emph{Pix3D}
and \emph{CUB-Birds-200-2011} we directly use the silhouette masks
provided in the database (but we do not use the additional labels). In
addition, for \emph{CUB-Birds-200-2011}, we also remove close-up views
and occluded images as their silhouettes do not provide complete shape
information, as well as images with birds with open wings due to the
scarcity of such images.  For the images from other sources, we
extract the silhouettes with an off-the-shelf segmentation
network~\cite{he2017mask} and manually remove instances with corrupted
masks.

\paragraph{Implementation Details}
We implemented our multi-projection shape GAN framework in
TensorFlow~\cite{Abadi:2016:TSL}.  For all experiments, the resulting
networks are trained with the ADAM optimizer with a $10^{-4}$ learning
rate, $\beta_1 = 0.5$, $\beta_2 = 0.9$, a $1:1$ training ratio between
the generator and the discriminators, and a batch size of $32$.  For
each training iteration, we generate one batch of voxel shapes, and for
each generated shape and projection discriminator compute a silhouette
image with randomly sampled viewpoint ($\sim \ZZ_i$). In addition, we
also sample one batch from the training set corresponding to each
projection.  During back-propagation, the gradients from each of the
discriminators are averaged to drive the generator training
(\autoref{eq:g}).

We limit viewpoint prediction in \emph{VP-MP-GAN} to azimuth angles,
since most collected images are dominated by viewpoint changes in the
azimuth angle. We split the azimuth range $[0, 2\pi]$ into $16$
uniformly distributed view-bins. For the view classifier training,
we randomly synthesize $10,\!000$ 3D shapes, and generate a silhouette
image for a random view within each of the view-bins, yielding a total
of $160,\!000$ training data pairs in each epoch. We cluster the view
distributions in $8$ clusters for all experiments and store the view
distributions $\ZZ_i$ for each discriminator in histograms. To avoid
outliers, we remove all bins with a probability less than $10\%$ and
renormalize the distribution.

Due to the intrinsic ambiguity of silhouette images for many
viewpoints (e.g., front and back silhouettes look the same), the
generated shapes may not align. Although the multi-projection GAN can
still learn the 3D shape distribution without alignment, such
ambiguities make the silhouette images from different views less
distinct, thus reducing the effectiveness of the multiple
discriminators. For datasets with known symmetry, we can leverage this
prior knowledge by explicitly modeling the symmetry.  In practice, we
enforce symmetry by only generating half of the voxel shape and
mirroring the remaining half over the symmetry axis. In our
experiments, we enforce symmetry for the chair and car dataset; but not
 for the bird dataset which exhibits non-symmetric poses.

Training \emph{VP-MP-GAN} at a resolution of $64^3$ takes on average
$40$ hours on $4$ Nvidia GTX 1080Ti cards.


\section{Experiments}
\label{sec:results}

To validate our 3D shape generator, we perform an ablation study to
demonstrate the impact of the number of projections and view clusters
(\autoref{sec:results_ablation}). In addition, we perform a comparison
against three related methods
(\autoref{sec:results_comparison}). Finally, we show that our solution
works well on non-synthetic image collections
(\autoref{sec:additional_results}).

\subsection{Ablation Study}
\label{sec:results_ablation}
We perform our ablation study on the synthetic chair dataset
(\autoref{sec:shape_implementation}) for which we also have reference
3D shapes (not used for training).  In this study, we evaluate the
quality of the generated results quantitatively using the FID
score~\cite{heusel2017gans} with an existing voxel classification
network~\cite{maturana2015voxnet} trained on the ShapeNet
dataset~\cite{Chang:2015:SNA} as the feature extractor.

\paragraph{Impact of Number of Projections}
To analyze the quality of the generator, we train \emph{MP-GAN} on a
pre-defined number of view distributions and pre-assign the training
images to the correct view clusters created by merging nearby
view-bins. \autoref{tab:md_range} summarizes the FID scores of MP-GAN
for a varying number of projections (and thus discriminators). Note
that MP-GAN reverts to a regular single projection GAN for the single
discriminator case. We observe that the FID score decreases, and thus
the generator quality improves, as the number of projections
increases.  However, we also observe a diminishing return (e.g., at
$16$ and $24$ projections) when increasing number of projections for a
fixed number of training data as the differences between the
projection distributions decreases.

\begin{table}[t]
\caption{FID score~\cite{heusel2017gans} of \emph{MP-GAN}
  trained on synthetic training data of chairs with reference
  viewpoint estimates for varying numbers of projections.}
\label{tab:md_range}
\resizebox{\linewidth}{!}{%
\begin{tabular}{llllllll}
\hline
\multicolumn{1}{|c|}{\begin{tabular}[c]{@{}c@{}}Num. of\\   Discriminators\end{tabular}} & \multicolumn{1}{c|}{1}      & \multicolumn{1}{c|}{2}      & \multicolumn{1}{c|}{4}     & \multicolumn{1}{c|}{6}     & \multicolumn{1}{c|}{8}     & \multicolumn{1}{c|}{16}	& \multicolumn{1}{c|}{24}\\ \hline
\multicolumn{1}{|c|}{FID Score}                                                         & \multicolumn{1}{c|}{79.61} & \multicolumn{1}{c|}{49.93} & \multicolumn{1}{c|}{36.22} & \multicolumn{1}{c|}{34.22} & \multicolumn{1}{c|}{33.27} & \multicolumn{1}{c|}{32.45}	& \multicolumn{1}{c|}{29.45}\\ \hline
                                                                                        &                             &                             &                            &                            &                            &	&\\
                                                                                        &                             &                             &                            &                            &	&	&                           
\end{tabular}
}
\vspace{-0.7cm}
\end{table}

\paragraph{Impact of Number of View Clusters}
We repeat the above experiment, but this time on \emph{VP-MP-GAN}
(i.e., with view prediction) on unannotated training silhouette
images. \autoref{tab:md_cluster} lists the FID scores for a varying
number of view clusters, which also changes the number of projections
and discriminators (i.e., each projection is assigned to a view
cluster).  The upper-bound for the FID scores is set by the
\emph{MP-GAN} (\autoref{tab:md_range}) as these are trained with the
exact viewpoint.  Compared to the upper-bound, we can see that the
scores for \emph{VP-MP-GAN} are similar or slightly larger. As the
number of view clusters increases, the inevitable inaccuracies
introduced by the view prediction weigh more on the accuracy,
resulting in a slightly larger FID score.

\begin{table}[t]
\caption{FID score for \emph{VP-MP-GAN} trained on the
  synthetic training data of chairs (with unknown viewpoints) for
  varying number of view clusters.}
\label{tab:md_cluster}
\resizebox{\linewidth}{!}{%
\begin{tabular}{llllllll}
\hline
\multicolumn{1}{|c|}{\begin{tabular}[c]{@{}c@{}}Num. of\\   Clusters\end{tabular}} & \multicolumn{1}{c|}{1}      & \multicolumn{1}{c|}{2}      & \multicolumn{1}{c|}{4}     & \multicolumn{1}{c|}{6}     & \multicolumn{1}{c|}{8}  & \multicolumn{1}{c|}{16} & \multicolumn{1}{c|}{24}   \\ \hline
\multicolumn{1}{|c|}{FID Score}                                                         & \multicolumn{1}{c|}{79.61} & \multicolumn{1}{c|}{53.83} & \multicolumn{1}{c|}{39.23} & \multicolumn{1}{c|}{35.22} & \multicolumn{1}{c|}{34.32} & \multicolumn{1}{c|}{34.10} & \multicolumn{1}{c|}{33.95}\\ \hline
                                                                                        &                             &                             &                            &                            &                           & & \\
                                                                                        &                             &                             &                            &                            &                           & &
\end{tabular}
}
\vspace{-0.7cm}
\end{table}

As detailed in~\autoref{sec:shape_view_estimation}, we iteratively
refine the viewpoint classifier (on $16$ pre-defined bins).
\autoref{tab:acc_iter} shows the improvement in view prediction
accuracy with each joint training iteration for \emph{VP-MP-GAN} with
$8$ view clusters.  The improvement in view prediction accuracy does
not only indicate that the view prediction improves, but also that the
learned shape distribution is closer to the target
distribution. \autoref{tab:acc} further demonstrates this by listing
the accuracy for varying number of view clusters (for $5$ iterations);
more view clusters result in a more accurate generator, which in turn
yields a more accurate view prediction.

\begin{table}[t]
\caption{Evolution of view classification accuracy during training of
  \emph{VP-MP-GAN} for $8$ view clusters.  Each (alternating)
  iteration includes $40,\!000$ GAN training iterations, and
  $40,\!000$ view predictor training iterations. The ``reference''
  column refers to the view predictor accuracy trained on silhouette
  images with exact viewpoint estimates.}
 \label{tab:acc_iter}
\resizebox{\linewidth}{!}{%
\begin{tabular}{lllllll}
\hline
\multicolumn{1}{|c|}{\begin{tabular}[c]{@{}c@{}}Num. of \\ Iterations\end{tabular}} & \multicolumn{1}{c|}{Ref.} & \multicolumn{1}{c|}{1}      & \multicolumn{1}{c|}{2}      & \multicolumn{1}{c|}{3}      & \multicolumn{1}{c|}{4}      & \multicolumn{1}{c|}{5}      \\ \hline
\multicolumn{1}{|c|}{Accuracy}                                                      & \multicolumn{1}{c|}{83.2\%}       & \multicolumn{1}{c|}{42.7\%} & \multicolumn{1}{c|}{66.5\%} & \multicolumn{1}{c|}{69.9\%} & \multicolumn{1}{c|}{73.3\%} & \multicolumn{1}{c|}{75.6\%} \\ \hline
                                                                                    &                                   &                             &                             &                             &                             &                             \\
                                                                                    &                                   &                             &                             &                             &                             &                            
\end{tabular}%
}
\vspace{-0.7cm}
\end{table}


\begin{table}[t]
\caption{View classification accuracy of \emph{VP-MP-GAN} for a
  varying number of view clusters after $5$ (alternating)
  iterations. The ``reference'' column refers to the view predictor
  accuracy trained on silhouette images with exact viewpoint
  estimates.}
\label{tab:acc}
\resizebox{\linewidth}{!}{%
\begin{tabular}{lllllll}
\hline
\multicolumn{1}{|c|}{\begin{tabular}[c]{@{}c@{}}Num. of \\ Clusters\end{tabular}} & \multicolumn{1}{c|}{Ref.} & \multicolumn{1}{c|}{1}      & \multicolumn{1}{c|}{2}      & \multicolumn{1}{c|}{4}      & \multicolumn{1}{c|}{6}      & \multicolumn{1}{c|}{8}      \\ \hline
\multicolumn{1}{|c|}{Accuracy}                                                      & \multicolumn{1}{c|}{83.2\%}       & \multicolumn{1}{c|}{42.7\%} & \multicolumn{1}{c|}{52.1\%} & \multicolumn{1}{c|}{54.5\%} & \multicolumn{1}{c|}{66.1\%} & \multicolumn{1}{c|}{75.6\%} \\ \hline
                                                                                    &                                   &                             &                             &                             &                             &                             \\
                                                                                    &                                   &                             &                             &                             &                             &                            
\end{tabular}%
}
\vspace{-0.7cm}
\end{table}

\begin{figure}[t]
	\begin{center}
	\begin{minipage}[t] {\linewidth}
	\centering
	\includegraphics[width=\textwidth]{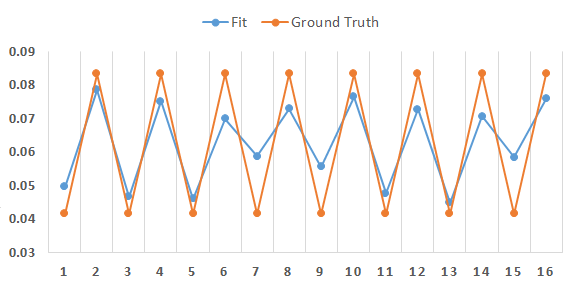}
	\end{minipage}
		\caption{Accuracy of the estimated view distribution
                  (blue) compared to a reference non-uniform
                  distribution of viewpoints (orange) for the
                  synthesized chair dataset.}
		\label{fig:distribution}
	\end{center}
\end{figure}

\begin{figure}[t]
	\begin{center}
		\begin{minipage}[t]{0.95\linewidth}
			\centering
			\includegraphics[width=\textwidth]{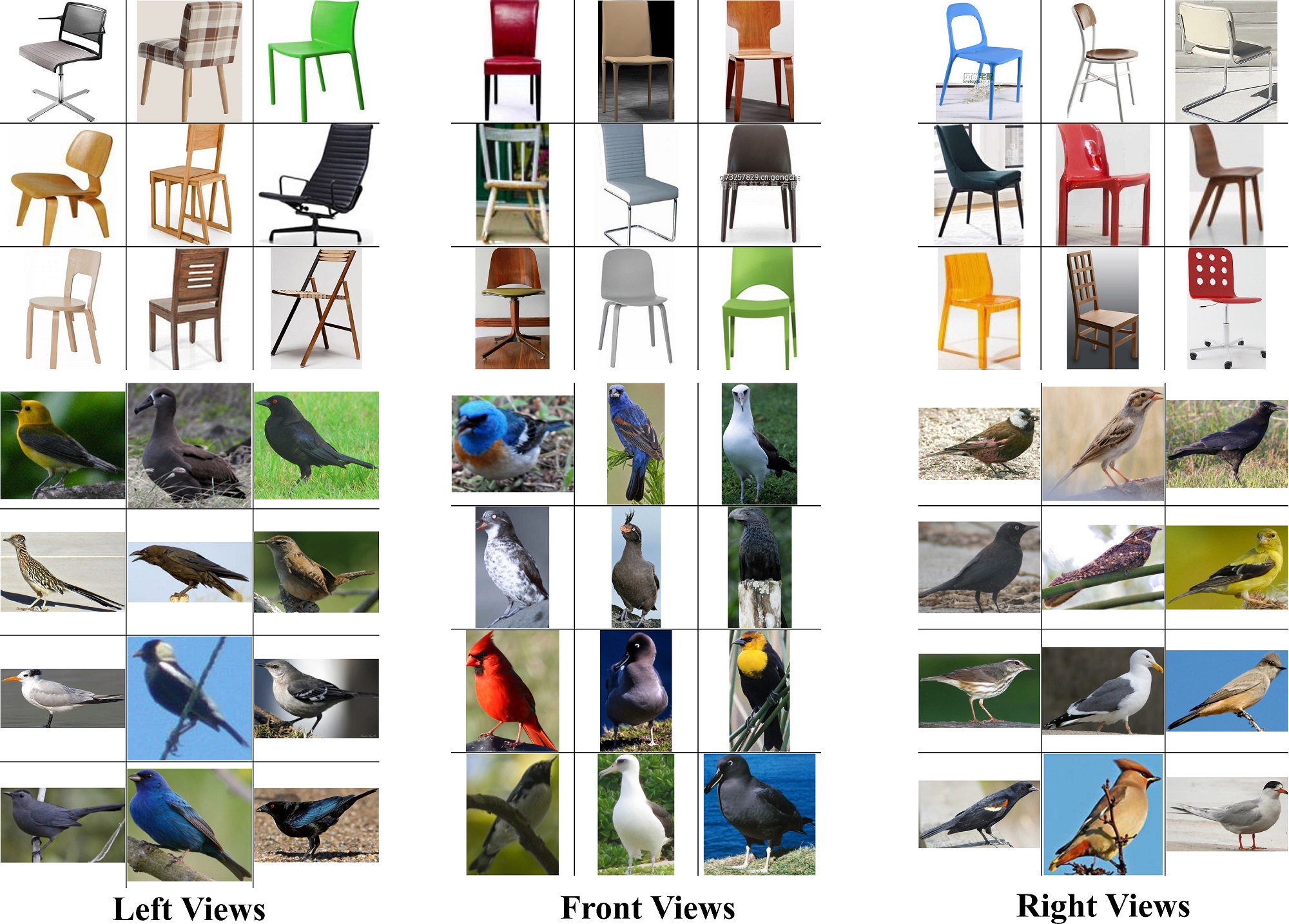}
		\end{minipage}
		\caption{Selection of view-classified training images
                  for the unannotated real-world chair and bird image
                  datasets.}
		\label{fig:view_real}
	\end{center}
\end{figure}

Finally, \autoref{fig:distribution} illustrates the view prediction
distribution accuracy on the synthetic chair dataset. Note that
\emph{VP-MP-GAN} is able to learn the correct distribution with
non-uniformly distributed views focused at $8$ peaks of the $16$
bins. For images gathered from internet repositories, our view
prediction also produces plausible results; \autoref{fig:view_real}
shows viewpoint classification results for the real-world chair and
bird datasets for selected views and images.

\subsection{Comparison}
\label{sec:results_comparison}

\begin{figure}[t]
\begin{center}
    \begin{minipage}[t] {\linewidth}
        \centering
        \includegraphics[width=\linewidth]{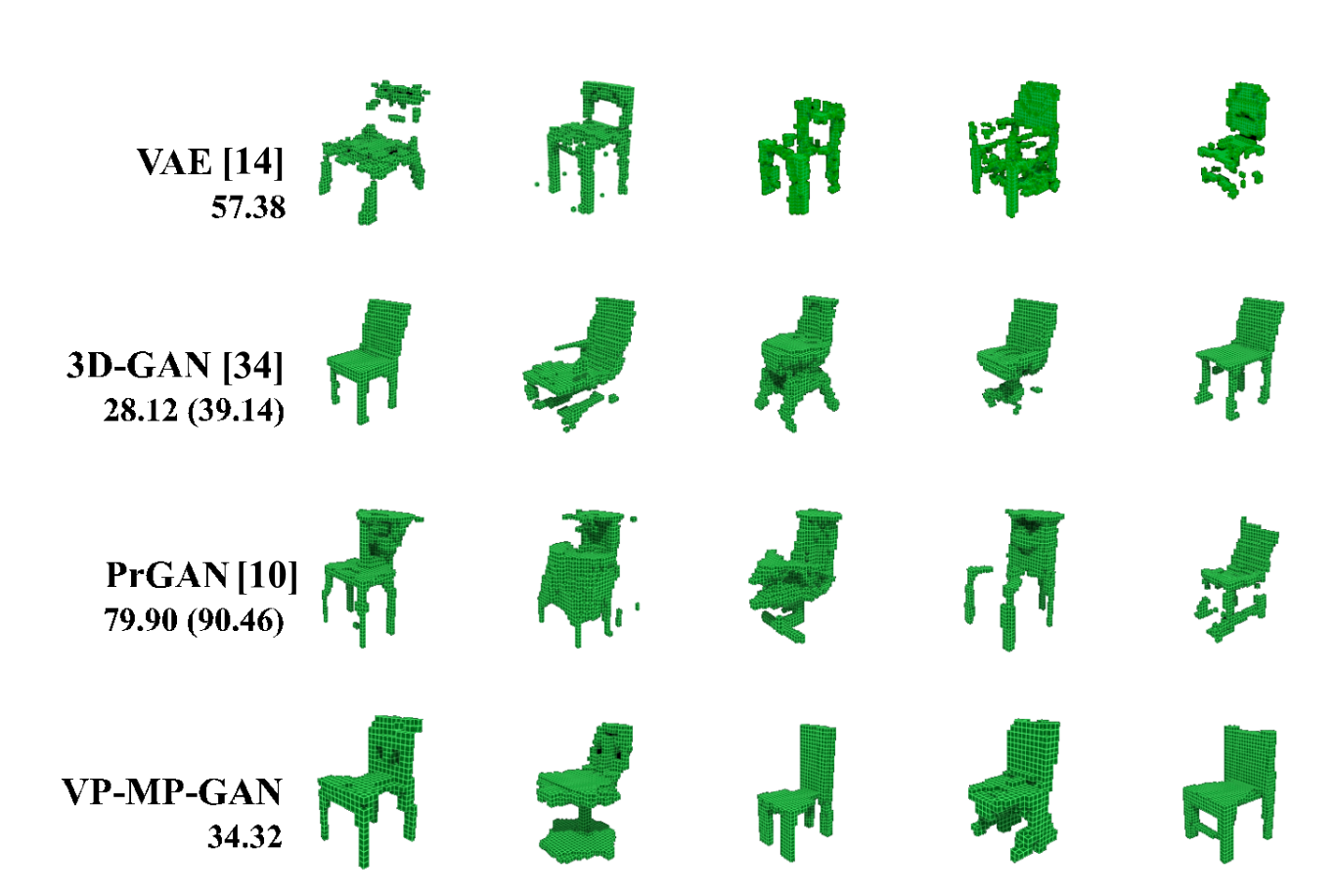}
        \vspace*{-0.1in}
        
    \end{minipage}
    \caption{Quantitative and qualitative comparison of four
      generators: VAE-based generator~\cite{henderson18bmvc} (top),
      3D-GAN~\cite{3dgan} (2nd row), PrGAN~\cite{Gadelha2017} (3rd
      row), and \emph{VP-MP-GAN} (bottom). The corresponding FID
      scores are shown on the left. We also report the FID score of
      the authors' original implementation in parenthesis.}
    \label{fig:voxel_compare}
\end{center}
\vspace{-0.7cm}
\end{figure}

\autoref{fig:voxel_compare} compares the results of our
\emph{VP-MP-GAN} to \emph{3D-GAN}~\cite{3dgan} and
\emph{PrGAN}~\cite{Gadelha2017} on the synthetic chair dataset. For a
fair comparison, we apply the symmetry constraint to both methods and
train both networks with our training data. We also list the FID score
(in parentheses) of authors' original implementations for reference.
\emph{3D-GAN} is directly trained on the reference 3D data, and
therefore scores a slightly higher FID score. Nevertheless, our
generated shapes exhibit a similar visual quality. Similar to our
method, \emph{PrGAN} is also trained on silhouette images without the
reference 3D data. However, \emph{PrGAN} assumes known viewpoints and
relies on a single discriminator only, resulting in a less accurate
shape generator.

\begin{figure}[t]
	\begin{center}
	  \centering
	  \includegraphics[width=0.95\linewidth]{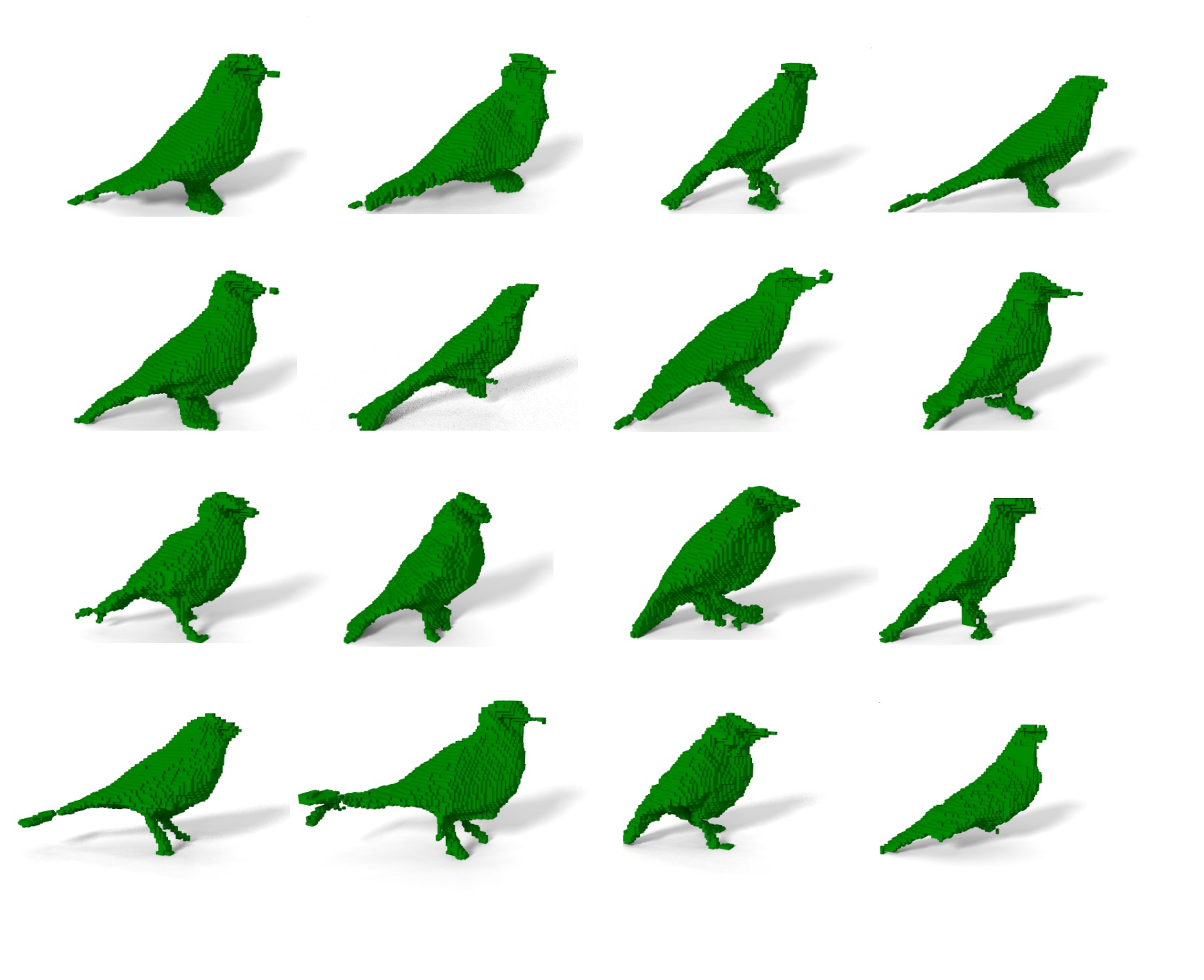}
	  \caption{Results generated by \emph{VP-MP-GAN} trained on the \emph{bird} dataset.}
	  \label{fig:voxel_bird}
	\end{center}
\vspace{-0.2cm}
\end{figure}

\begin{figure}[t]
	\begin{center}
	  \centering
	  \includegraphics[width=0.95\linewidth]{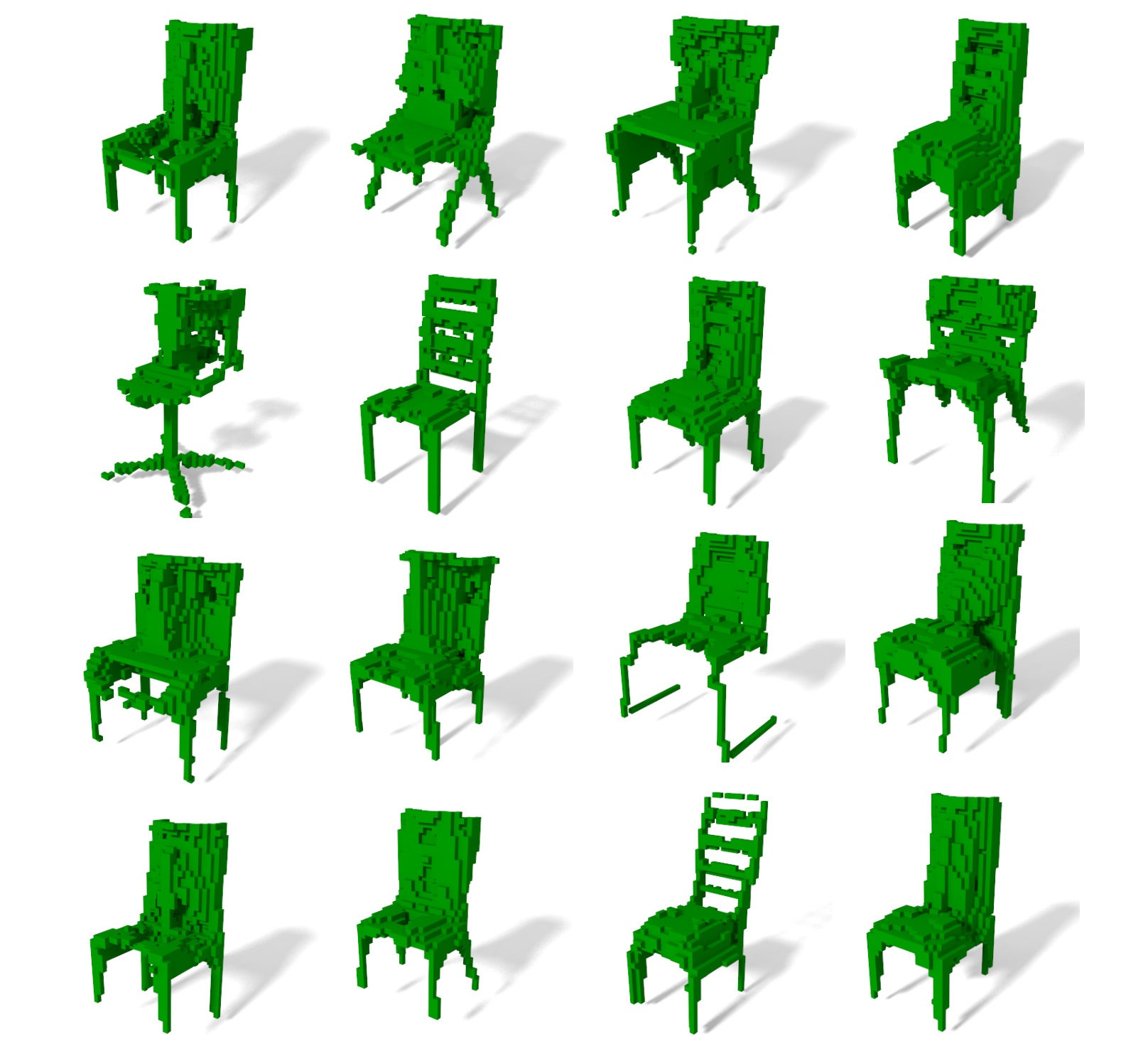}
	  \caption{Results generated by \emph{VP-MP-GAN} trained on
            the internet-mined \emph{chair} dataset.}
	  \label{fig:voxel_chair}
	\end{center}
\vspace{-0.7cm}
\end{figure}

\begin{figure}[t]
	\begin{center}
	  \includegraphics[width=0.95\linewidth]{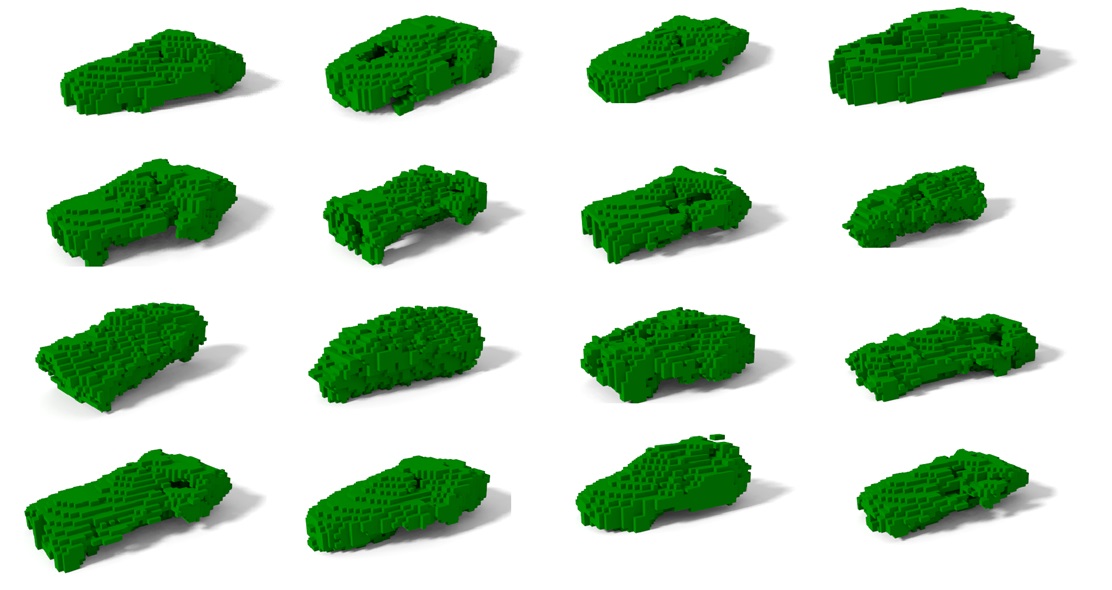}
	  \caption{Results generated by \emph{VP-MP-GAN} trained on the \emph{car} dataset.}
	  \label{fig:voxel_car}
	\end{center}
\vspace{-0.2cm}
\end{figure}

Recently, Henderson~\etal~\cite{henderson18bmvc} introduced a method
for learning 3D shape distributions from shaded images using a VAE
approach. Their method uses a 3D mesh representation instead of a
voxel grid, making a direct comparison difficult.  We therefore, adapt
and retrain their method with a voxel generator instead of a mesh
generator on silhouette images with a uniform view distribution and
added symmetry constraint. \autoref{fig:voxel_compare} shows that
\emph{VP-MP-GAN} produces higher quality voxel shapes and exhibits a
lower FID score on the synthetic chair dataset.

\newcommand{\FigureSizeMDHD}{0.19}
\begin{figure*}[t]
	\begin{center}
		\begin{minipage}[t] {\linewidth}
		\begin{center}
		\begin{minipage}[t] {\FigureSizeMDHD\textwidth}
			\centering
			\includegraphics[width=\textwidth]{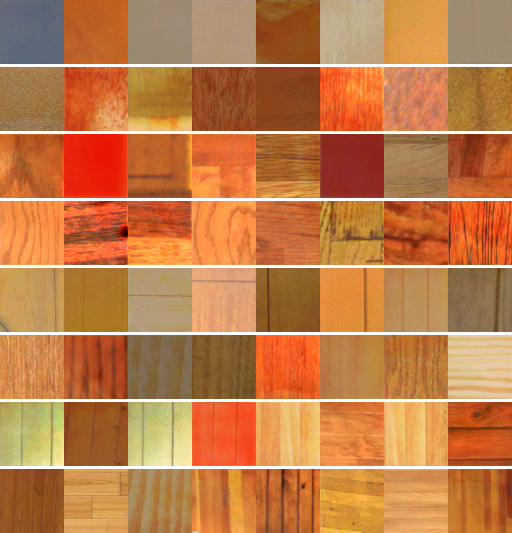}
			\vspace*{-0.1in}
		\end{minipage}
		\begin{minipage}[t] {\FigureSizeMDHD\textwidth}
			\centering
			\includegraphics[width=\textwidth]{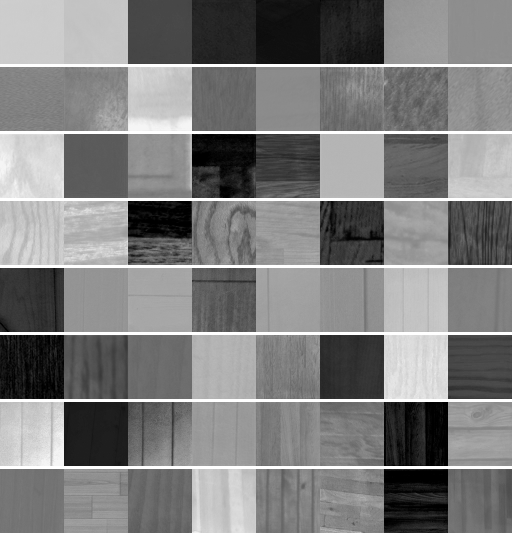}
		\end{minipage}
		\begin{minipage}[t] {\FigureSizeMDHD\textwidth}
			\centering
			\includegraphics[width=\textwidth]{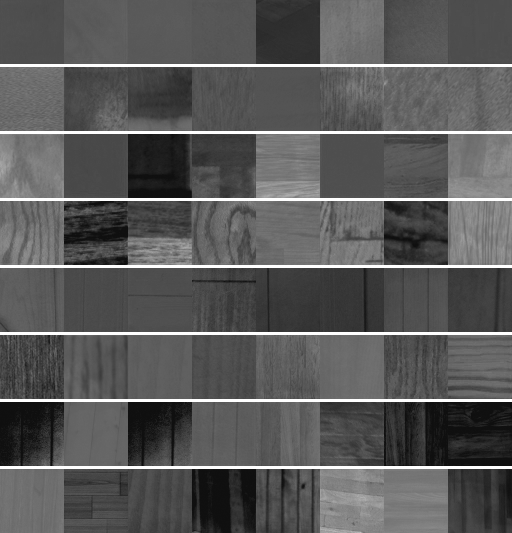}
		\end{minipage}
		\begin{minipage}[t] {\FigureSizeMDHD\textwidth}
			\centering
			\includegraphics[width=\textwidth]{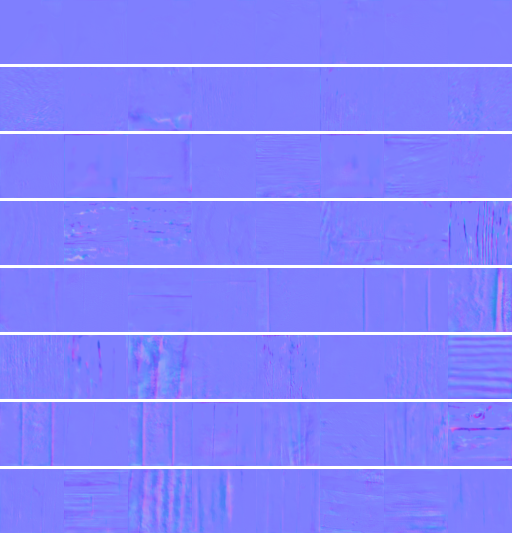}
		\end{minipage}
		\begin{minipage}[t] {\FigureSizeMDHD\textwidth}
			\centering
			\includegraphics[width=\textwidth]{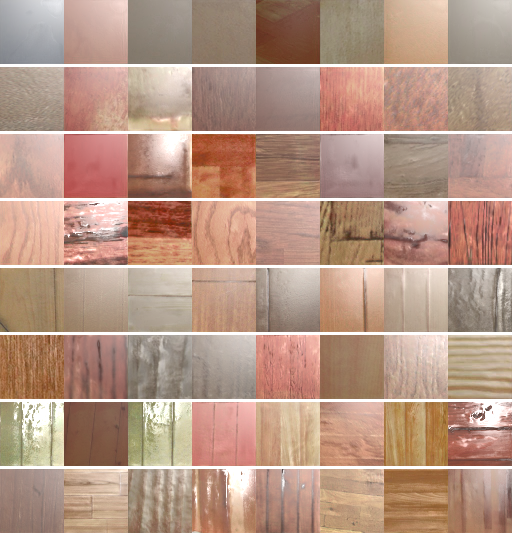}
		\end{minipage}

		\begin{minipage}[t] {\FigureSizeMDHD\textwidth}
			\centering
			\includegraphics[width=\textwidth]{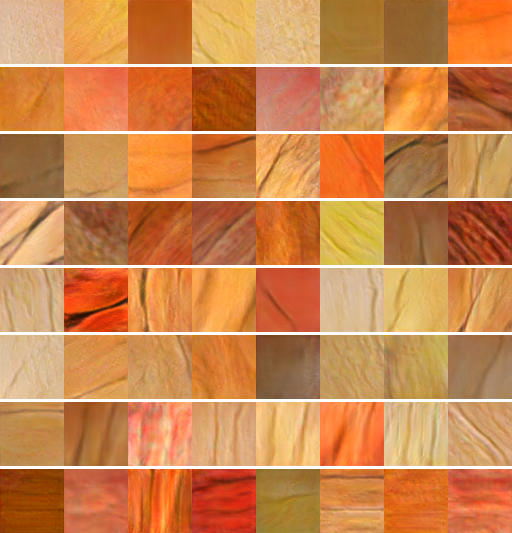}
			{Diffuse}
		\end{minipage}
		\begin{minipage}[t] {\FigureSizeMDHD\textwidth}
			\centering
			\includegraphics[width=\textwidth]{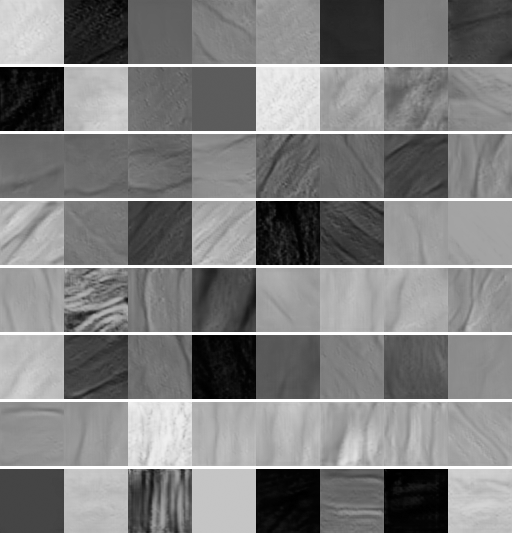}
			{Specular}
		\end{minipage}
		\begin{minipage}[t] {\FigureSizeMDHD\textwidth}
			\centering
			\includegraphics[width=\textwidth]{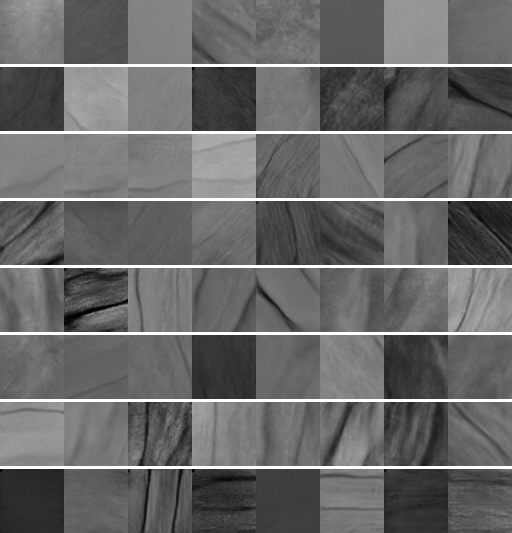}
			{Roughness}
		\end{minipage}
		\begin{minipage}[t] {\FigureSizeMDHD\textwidth}
			\centering
			\includegraphics[width=\textwidth]{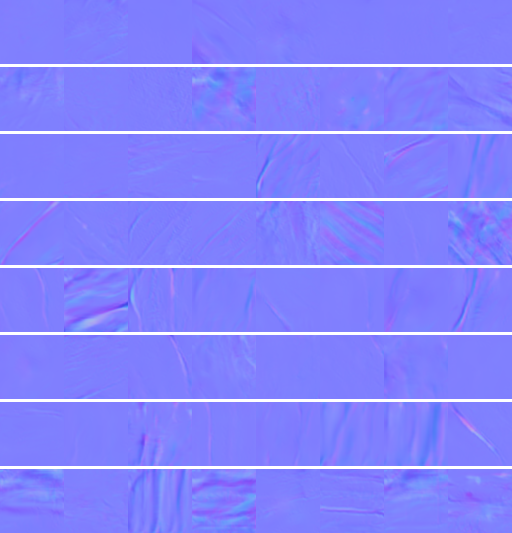}
			{Normal}
		\end{minipage}
		\begin{minipage}[t] {\FigureSizeMDHD\textwidth}
			\centering
			\includegraphics[width=\textwidth]{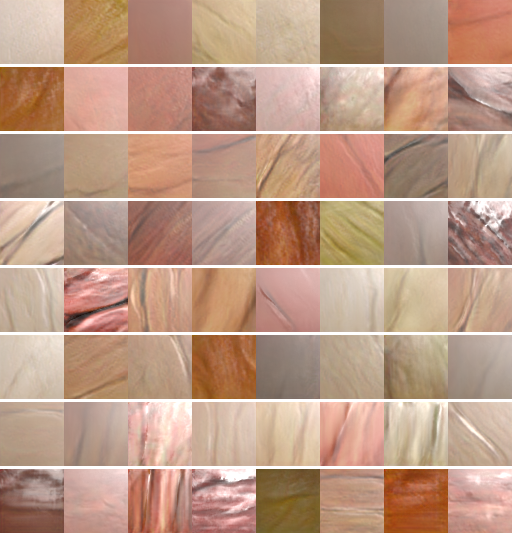}
			{Rendering}
		\end{minipage}
		\end{center}
		\end{minipage}
		\caption{A selection of SVBRDFs generated with our
                  multi-projection GAN (bottom) compared to samples
                  drawn from the training dataset (top). }
		\label{fig:svbrdf_result}
	\end{center}
\vspace{-0.5cm}
\end{figure*}

\subsection{Results on Real-world Datasets}
\label{sec:additional_results}
We demonstrate our method's ability to learn the 3D shape distribution
from real-world image datasets on three real-world image
collections. \autoref{fig:voxel_bird} shows examples of generated bird
shapes. Note that there currently does not exist a database of 3D bird
shapes, and thus the only way to learn a bird-shape generator is
directly from images.  \autoref{fig:voxel_chair}
and~\autoref{fig:voxel_car} show results of generated chairs and cars,
respectively, trained from photographs mined from online
photo-collections (without using any synthetic images generated from
ShapeNet).  As can be seen, in all cases, \emph{VP-MP-GAN} is able to
produce high quality voxel shapes from unannotated silhouette images.
We also refer to the supplemental material for more results on these
datasets.

\paragraph{Limitations:} 
\emph{VP-MP-GAN} infers the distribution of 3D shapes from unoccluded
silhouette images.  Similar to classic computer vision methods that
infer a single shape from silhouette images, our method can also not
model concavities, and it requires a complete unoccluded view of the
objects in the images.  Using depth images instead of silhouette
images can resolve the concavity issue. However, we deliberately did
not go this route as depth images are not readily available and need
to be specially captured.


\section{SVBRDF Modeling}
\label{sec:results_svbrdf}
Our multi-projection GAN framework is not restricted to 3D shape
modeling only, and it can potentially be applied to other applications
that model high-dimensional data for which easy access to
low-dimensional projections is available.  We demonstrate the
generality of the multi-projection framework by learning the
distribution of spatially-varying material appearance in the form of
property maps for spatially-varying bidirectional reflectance
distribution functions (SVBRDF) for certain kinds of natural materials
such as wood, metal, and plastics.  A SVBRDF is a 6D function that
describes how the appearance varies with position, view angle, and
incident lighting direction.  We model the SVBRDF by a set of four 2D
parameter maps that describe the Cook-Torrance BRDF
model's~\cite{Cook:1982:ARM} reflectance parameters for each surface
point (i.e., diffuse albedo, specular albedo, specular roughness, and
surface normal).

Unlike textures which can be acquired with a single photograph,
accurate characterization of the full SVBRDF and surface normal
details of a material is a complex and time-consuming
process~\cite{Weinmann:2015:AGR}.  However, it is relatively easy to
collect uncorresponded example maps of certain parameters (e.g.,
photometric stereo for normal maps, diffuse albedo via cross-polarized
photographs, etc).  Furthermore, it is easy to obtain photographs of
materials under unknown lighting from internet photo-collections.
Both of these types of images represent projections of the 6D SVBRDF.
An image of a property map (diffuse or specular albedo, specular
roughness, or normal maps) corresponds to a trivial projection that
\emph{``selects''} a single property map.  A photograph under unknown
lighting corresponds to a \emph{``rendering''} projection where the
unknown natural lighting distribution is modeled by the latent
projection parameter $\ZZ$.

\autoref{fig:svbrdf_result} shows (corresponded) generated appearance
property maps obtained with our multi-projection GAN, trained on a
dataset containing partial appearance property maps as well as real
photographs from the OpenSurfaces dataset~\cite{Bell2013}.  The last
column in~\autoref{fig:svbrdf_result} shows a rendering of the
generated material under a novel lighting condition.  As can be seen,
the generated materials produce a plausible \emph{wood} material
appearance. We refer to the supplemental material for additional
SVBRDF results, a didactic example using the MNIST dataset, as well as
the technical details including the training process, network
structure, and the collection and preparation of the training data.

\section{Conclusion}
We proposed a novel weakly supervised method for learning the
distribution of 3D shapes for a class of objects from unoccluded
silhouette images.  Key to our method is a novel multi-projection
formulation of GANs that learns a high-dimensional distribution (i.e.,
voxel grid) from multiple, easier to acquire, lower-dimensional
training data consisting of silhouette images from different objects
from multiple viewpoints. Our method does not require that the
silhouettes from multiple views are corresponded, nor that the
viewpoints are known.  The generator network is trained with cues from
multiple discriminators in parallel. Each discriminator operates on
the subset of the training data corresponding to a particular
viewpoint.  Our second contribution is a novel joint training strategy
for training the view prediction network in an iterative fashion with
the multi-projection GAN.  We demonstrated the effectiveness of our 3D
voxel generator on both synthetic and real-world
datasets. Furthermore, we showed that our multi-projection framework
is more generally applicable than to 3D shape modeling only, and
demonstrated this by training a SVBRDF generator from 2D images.

\paragraph{Acknowledgments} We would like to thank the reviewers for 
their constructive feedback. We also thank Baining Guo for discussions
and suggestions. Pieter Peers was partially supported by 
NSF grant IIS-1350323 and gifts from Google, Activision, and Nvidia.

{\small
\bibliographystyle{ieee}
\bibliography{paper}

\begin{thebibliography}{10}\itemsep=-1pt

\bibitem{Abadi:2016:TSL}
M.~Abadi, P.~Barham, J.~Chen, Z.~Chen, A.~Davis, J.~Dean, M.~Devin,
  S.~Ghemawat, G.~Irving, M.~Isard, M.~Kudlur, J.~Levenberg, R.~Monga,
  S.~Moore, D.~G. Murray, B.~Steiner, P.~Tucker, V.~Vasudevan, P.~Warden,
  M.~Wicke, Y.~Yu, and X.~Zheng.
\newblock Tensorflow: A system for large-scale machine learning.
\newblock In {\em OSDI}, pages 265--283, 2016.

\bibitem{Arjovsky:2017:WGAN}
M.~Arjovsky, S.~Chintala, and L.~Bottou.
\newblock Wasserstein generative adversarial networks.
\newblock In {\em ICML}, pages 214--223, 2017.

\bibitem{Bell2013}
S.~Bell, P.~Upchurch, N.~Snavely, and K.~Bala.
\newblock Open{S}urfaces: A richly annotated catalog of surface appearance.
\newblock {\em ACM Trans. on Graph.}, 32(4), 2013.

\bibitem{bora2018ambientgan}
A.~Bora, E.~Price, and A.~G. Dimakis.
\newblock Ambientgan: Generative models from lossy measurements.
\newblock In {\em ICLR}, 2018.

\bibitem{Chang:2015:SNA}
A.~X. Chang, T.~A. Funkhouser, L.~J. Guibas, P.~Hanrahan, Q.~Huang, Z.~Li,
  S.~Savarese, M.~Savva, S.~Song, H.~Su, J.~Xiao, L.~Yi, and F.~Yu.
\newblock Shapenet: An information-rich 3d model repository.
\newblock {\em arXiv}, 2015.

\bibitem{choy20163d}
C.~B. Choy, D.~Xu, J.~Gwak, K.~Chen, and S.~Savarese.
\newblock 3d-r2n2: A unified approach for single and multi-view 3d object
  reconstruction.
\newblock In {\em ECCV}, 2016.

\bibitem{Cook:1982:ARM}
R.~L. Cook and K.~E. Torrance.
\newblock A reflectance model for computer graphics.
\newblock {\em ACM Trans. Graph.}, 1(1):7--24, 1982.

\bibitem{Durugkar:2016:GMA}
I.~P. Durugkar, I.~Gemp, and S.~Mahadevan.
\newblock Generative multi-adversarial networks.
\newblock {\em arXiv}, 2016.

\bibitem{fan2017point}
H.~Fan, H.~Su, and L.~J. Guibas.
\newblock A point set generation network for 3d object reconstruction from a
  single image.
\newblock In {\em CVPR}, volume~2, page~6, 2017.

\bibitem{Gadelha2017}
M.~Gadelha, S.~Maji, and R.~Wang.
\newblock {3D Shape Induction from 2D Views of Multiple Objects}.
\newblock In {\em International Conference on 3D Vision}, dec 2017.

\bibitem{girdhar2016learning}
R.~Girdhar, D.~F. Fouhey, M.~Rodriguez, and A.~Gupta.
\newblock Learning a predictable and generative vector representation for
  objects.
\newblock In {\em ECCV}, pages 484--499, 2016.

\bibitem{Goodfellow2014}
I.~J. Goodfellow, J.~Pouget-Abadie, M.~Mirza, B.~Xu, D.~Warde-Farley, S.~Ozair,
  A.~Courville, and Y.~Bengio.
\newblock {Generative Adversarial Nets}.
\newblock In {\em NIPS}, 2014.

\bibitem{he2017mask}
K.~He, G.~Gkioxari, P.~Doll{\'a}r, and R.~Girshick.
\newblock Mask r-cnn.
\newblock In {\em ICCV}, pages 2980--2988, 2017.

\bibitem{henderson18bmvc}
P.~Henderson and V.~Ferrari.
\newblock Learning to generate and reconstruct 3d meshes with only 2d
  supervision.
\newblock In {\em BMVC}, 2018.

\bibitem{heusel2017gans}
M.~Heusel, H.~Ramsauer, T.~Unterthiner, B.~Nessler, and S.~Hochreiter.
\newblock Gans trained by a two time-scale update rule converge to a local nash
  equilibrium.
\newblock In {\em NIPS}, pages 6626--6637, 2017.

\bibitem{jiang2018gal}
L.~Jiang, S.~Shi, X.~Qi, and J.~Jia.
\newblock Gal: Geometric adversarial loss for single-view 3d-object
  reconstruction.
\newblock In {\em ECCV}, pages 820--834, 2018.

\bibitem{kanazawa2018end}
A.~Kanazawa, M.~J. Black, D.~W. Jacobs, and J.~Malik.
\newblock End-to-end recovery of human shape and pose.
\newblock In {\em CVPR}, pages 7122--7131, 2018.

\bibitem{cmrKanazawa18}
A.~Kanazawa, S.~Tulsiani, A.~A. Efros, and J.~Malik.
\newblock Learning category-specific mesh reconstruction from image
  collections.
\newblock In {\em ECCV}, 2018.

\bibitem{KrauseStarkDengFei-Fei_3DRR2013}
J.~Krause, M.~Stark, J.~Deng, and L.~Fei-Fei.
\newblock 3d object representations for fine-grained categorization.
\newblock In {\em 3dRR-13}, 2013.

\bibitem{Massa:2016:CMC}
F.~Massa, R.~Marlet, and M.~Aubry.
\newblock Crafting a multi-task {CNN} for viewpoint estimation.
\newblock In {\em BMVC}, 2016.

\bibitem{maturana2015voxnet}
D.~Maturana and S.~Scherer.
\newblock Voxnet: A 3d convolutional neural network for real-time object
  recognition.
\newblock In {\em IROS}, pages 922--928, 2015.

\bibitem{miyato2018spectral}
T.~Miyato, T.~Kataoka, M.~Koyama, and Y.~Yoshida.
\newblock Spectral normalization for generative adversarial networks.
\newblock In {\em ICLR}, 2018.

\bibitem{Neyshabur:2017:SGT}
B.~Neyshabur, S.~Bhojanapalli, and A.~Chakrabarti.
\newblock Stabilizing {GAN} training with multiple random projections.
\newblock {\em arXiv}, 2017.

\bibitem{Radford:2015:URL}
A.~Radford, L.~Metz, and S.~Chintala.
\newblock Unsupervised representation learning with deep convolutional
  generative adversarial networks.
\newblock {\em arXiv}, 2015.

\bibitem{sinha2017surfnet}
A.~Sinha, A.~Unmesh, Q.~Huang, and K.~Ramani.
\newblock Surfnet: Generating 3d shape surfaces using deep residual networks.
\newblock In {\em CVPR}, volume~1, 2017.

\bibitem{Su:2015:RCV}
H.~Su, C.~R. Qi, Y.~Li, and L.~J. Guibas.
\newblock Render for cnn: Viewpoint estimation in images using cnns trained
  with rendered 3d model views.
\newblock In {\em ICCV}, December 2015.

\bibitem{pix3d}
X.~Sun, J.~Wu, X.~Zhang, Z.~Zhang, C.~Zhang, T.~Xue, J.~B. Tenenbaum, and W.~T.
  Freeman.
\newblock Pix3d: Dataset and methods for single-image 3d shape modeling.
\newblock In {\em CVPR}, 2018.

\bibitem{mvcTulsiani18}
S.~Tulsiani, A.~A. Efros, and J.~Malik.
\newblock Multi-view consistency as supervisory signal for learning shape and
  pose prediction.
\newblock In {\em CVPR}, 2018.

\bibitem{Tulsiani:2017:MSS}
S.~Tulsiani, T.~Zhou, A.~A. Efros, and J.~Malik.
\newblock Multi-view supervision for single-view reconstruction via
  differentiable ray consistency.
\newblock In {\em CVPR}, 2017.

\bibitem{WahCUB_200_2011}
C.~Wah, S.~Branson, P.~Welinder, P.~Perona, and S.~Belongie.
\newblock {The Caltech-UCSD Birds-200-2011 Dataset}.
\newblock Technical Report CNS-TR-2011-001, California Institute of Technology,
  2011.

\bibitem{Weinmann:2015:AGR}
M.~Weinmann and R.~Klein.
\newblock Advances in geometry and reflectance acquisition.
\newblock In {\em ACM SIGGRAPH Asia, Course Notes}, 2015.

\bibitem{Wu:2017:MarrNet}
J.~Wu, Y.~Wang, T.~Xue, X.~Sun, B.~Freeman, and J.~Tenenbaum.
\newblock Marrnet: 3d shape reconstruction via 2.5d sketches.
\newblock In {\em NIPS}.

\bibitem{WU:2016:SII}
J.~Wu, T.~Xue, J.~J. Lim, Y.~Tian, J.~B. Tenenbaum, A.~Torralba, and W.~T.
  Freeman.
\newblock Single image 3d interpreter network.
\newblock In {\em ECCV}, 2016.

\bibitem{3dgan}
J.~Wu, C.~Zhang, T.~Xue, W.~T. Freeman, and J.~B. Tenenbaum.
\newblock Learning a probabilistic latent space of object shapes via 3d
  generative-adversarial modeling.
\newblock In {\em NIPS}, pages 82--90, 2016.

\bibitem{Yan:2016:PTN}
X.~Yan, J.~Yang, E.~Yumer, Y.~Guo, and H.~Lee.
\newblock Perspective transformer nets: Learning single-view 3d object
  reconstruction without 3d supervision.
\newblock In D.~D. Lee, M.~Sugiyama, U.~V. Luxburg, I.~Guyon, and R.~Garnett,
  editors, {\em NIPS}, pages 1696--1704. 2016.

\bibitem{Yang18}
B.~Yang, S.~Rosa, A.~Markham, N.~Trigoni, and H.~Wen.
\newblock Dense 3d object reconstruction from a single depth view.
\newblock In {\em TPAMI}, 2018.

\bibitem{zhu2017rethinking}
R.~Zhu, H.~K. Galoogahi, C.~Wang, and S.~Lucey.
\newblock Rethinking reprojection: Closing the loop for pose-aware shape
  reconstruction from a single image.
\newblock In {\em ICCV}, pages 57--65, 2017.

\end{thebibliography}
}

\end{document}